\documentclass[10pt,twocolumn,letterpaper]{article}

\usepackage[accsupp]{axessibility}
\usepackage{cvpr}              

\usepackage{graphicx}
\usepackage{amsmath}
\usepackage{amssymb}
\usepackage{booktabs}
\usepackage{color}
\usepackage{colortbl}
\usepackage{multirow}

\usepackage[pagebackref,breaklinks,colorlinks]{hyperref}

\usepackage[capitalize]{cleveref}
\crefname{section}{Sec.}{Secs.}
\Crefname{section}{Section}{Sections}
\Crefname{table}{Table}{Tables}
\crefname{table}{Tab.}{Tabs.}
\crefname{figure}{Fig.}{Figs.}
\Crefname{figure}{Figure}{Figures}
\crefname{equation}{Eq.}{Eqs.}
\Crefname{equation}{Equation}{Equations}

\newtheorem{remark}{Remark}

\def\cvprPaperID{6339} 
\def\confName{CVPR}
\def\confYear{2023}

\begin{document}

\title{Self-positioning Point-based Transformer for Point Cloud Understanding}

\author{Jinyoung Park$^1$\thanks{First two authors have equal contribution.} , Sanghyeok Lee$^{1*}$, Sihyeon Kim$^1$, Yunyang Xiong$^2$, Hyunwoo J. Kim$^1$\thanks{is the corresponding author.}\\
$^1$Korea University, $^2$Meta Reality Labs\\
{\tt\small \{lpmn678, cat0626, sh\_bs15, hyunwoojkim\}@korea.ac.kr}\\
{\tt\small yunyang@fb.com}
}
\maketitle
\newcommand{\Pointset}{\mathcal{P}}
\newcommand{\Pointgroup}{\mathcal{G}}
\newcommand{\Input}{p}
\newcommand{\Rb}{\mathbb{R}}
\newcommand{\Vpoint}{\delta}
\newcommand{\Vfeat}{\boldsymbol{\psi}}
\newcommand{\Zfeat}{\mathbf{z}}
\newcommand{\Feat}{\mathbf{f}}
\newcommand{\Cen}{x}
\newcommand{\PE}{\phi}
\newcommand{\Pc}{\mathcal{P}}
\newcommand{\Gc}{\mathcal{G}}
\newcommand{\FFN}{\text{MLP}}
\newcommand{\LN}{\text{LN}}
\newcommand{\shl}{\textcolor[rgb]{1,0,0}}
\newcommand{\jyp}{\textcolor{blue}}
\newcommand{\sk}{\textcolor[rgb]{0.5,0.5,0}}
\newcommand{\hjk}[1]{{\color[rgb]{0, 0, 0} #1}}
\newcommand{\name}{SPoTr}
\newcommand{\mycyan}[1]{{\color[rgb]{0.2800,0.4200,0.5000}#1}}

\definecolor{LightRed}{rgb}{1,0.8,0.8} 
\definecolor{LightYellow}{rgb}{1,1,0.8} 
\definecolor{LightBlue}{rgb}{0.8,0.9,1} 
\definecolor{LightOrange}{rgb}{1,0.9,0.8} 

\begin{abstract}
Transformers have shown superior performance on various computer vision tasks with their capabilities to capture long-range dependencies. 
Despite the success, it is challenging to directly apply Transformers on point clouds due to their quadratic cost in  the number of points.
In this paper, we present a \textbf{S}elf-\textbf{Po}sitioning point-based \textbf{Tr}ansformer~(SPoTr), which is designed to capture both local and global shape contexts with reduced complexity.
Specifically, this architecture consists of local self-attention and self-positioning point-based global cross-attention.
The self-positioning points, adaptively located based on the input shape, consider both spatial and semantic information with {disentangled attention} to improve expressive power.
With the self-positioning points, we propose a novel global cross-attention mechanism for point clouds, which improves the scalability of global self-attention by allowing the attention module to compute attention weights with only a small set of self-positioning points.
Experiments show the effectiveness of SPoTr on three point cloud tasks such as shape classification, part segmentation, and scene segmentation. 
In particular, our proposed model achieves an accuracy gain of 2.6\% over the previous best models on shape classification with ScanObjectNN.
We also provide qualitative analyses to demonstrate the interpretability of self-positioning points.
The code of SPoTr is available at \url{https://github.com/mlvlab/SPoTr}.

\end{abstract}

\section{Introduction}
    \label{sec:1}
Point clouds have been widely applied in various areas such as autonomous driving, robotics, and augmented reality.  
Since the point cloud is an unordered set of points with irregular structures, adopting convolutional neural networks~(CNNs) on point clouds is challenging.
Some works devoted effort to transforming point clouds into regular structures, such as projection to multi-view images~\cite{su2015multi,chen2017multi} and voxelization~\cite{maturana2015voxnet,zhou2018voxelnet}.
Others have tried to preserve the structure and design a convolution on the point space~\cite{liu2019relation,thomas2019kpconv,li2018pointcnn,atzmon2018point,xu2018spidercnn,wu2019pointconv,xu2021paconv}.
However, the ability to capture long-range dependencies is limited in most convolution-based approaches, while it is crucial to understand global shape context, especially with real-world data~\cite{uy2019revisiting}.

\begin{figure}[t]
     \centering
     \begin{subfigure}[b]{0.156\textwidth}
         \centering
         \includegraphics[trim=70 70 70 70,clip,width=\textwidth]{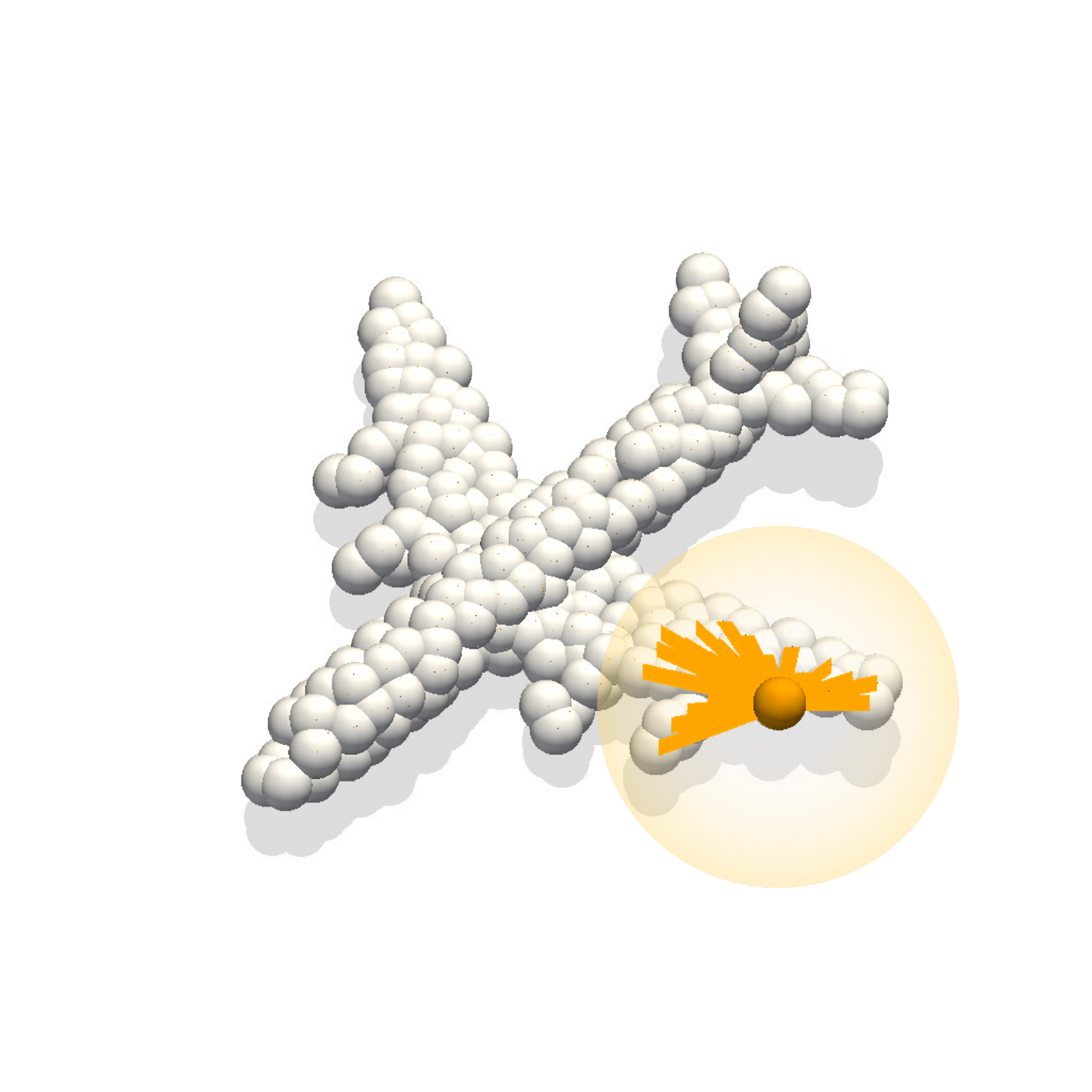}
         \caption{Local attention}
         \label{fig:local_att}
     \end{subfigure}
     \begin{subfigure}[b]{0.156\textwidth}
         \centering
         \includegraphics[trim=70 70 70 70,clip,width=\textwidth]{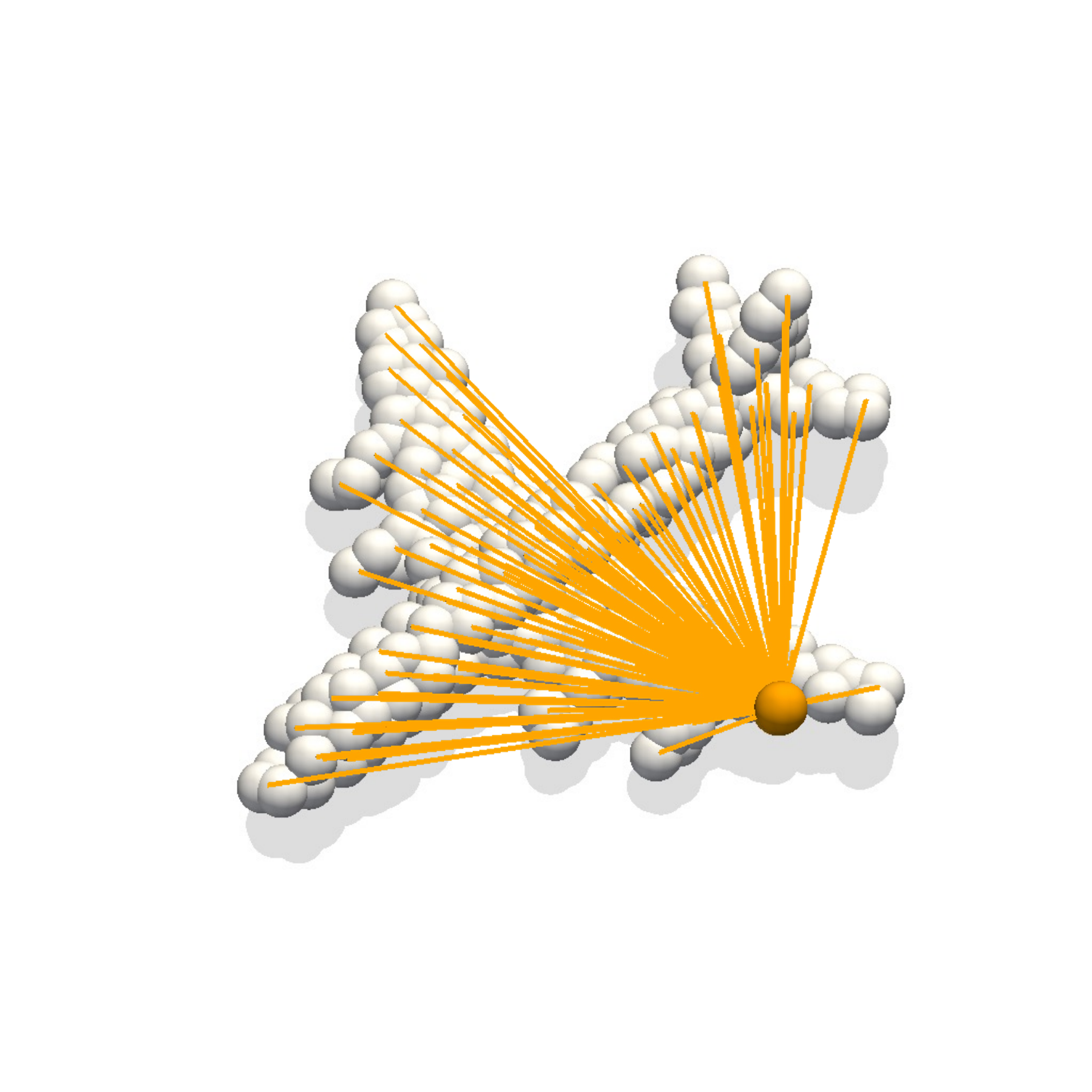}
         \caption{Global attention}
         \label{fig:global_att}
     \end{subfigure}
     \begin{subfigure}[b]{0.156\textwidth}
         \centering
         \includegraphics[trim=70 70 70 70,clip,width=\textwidth]{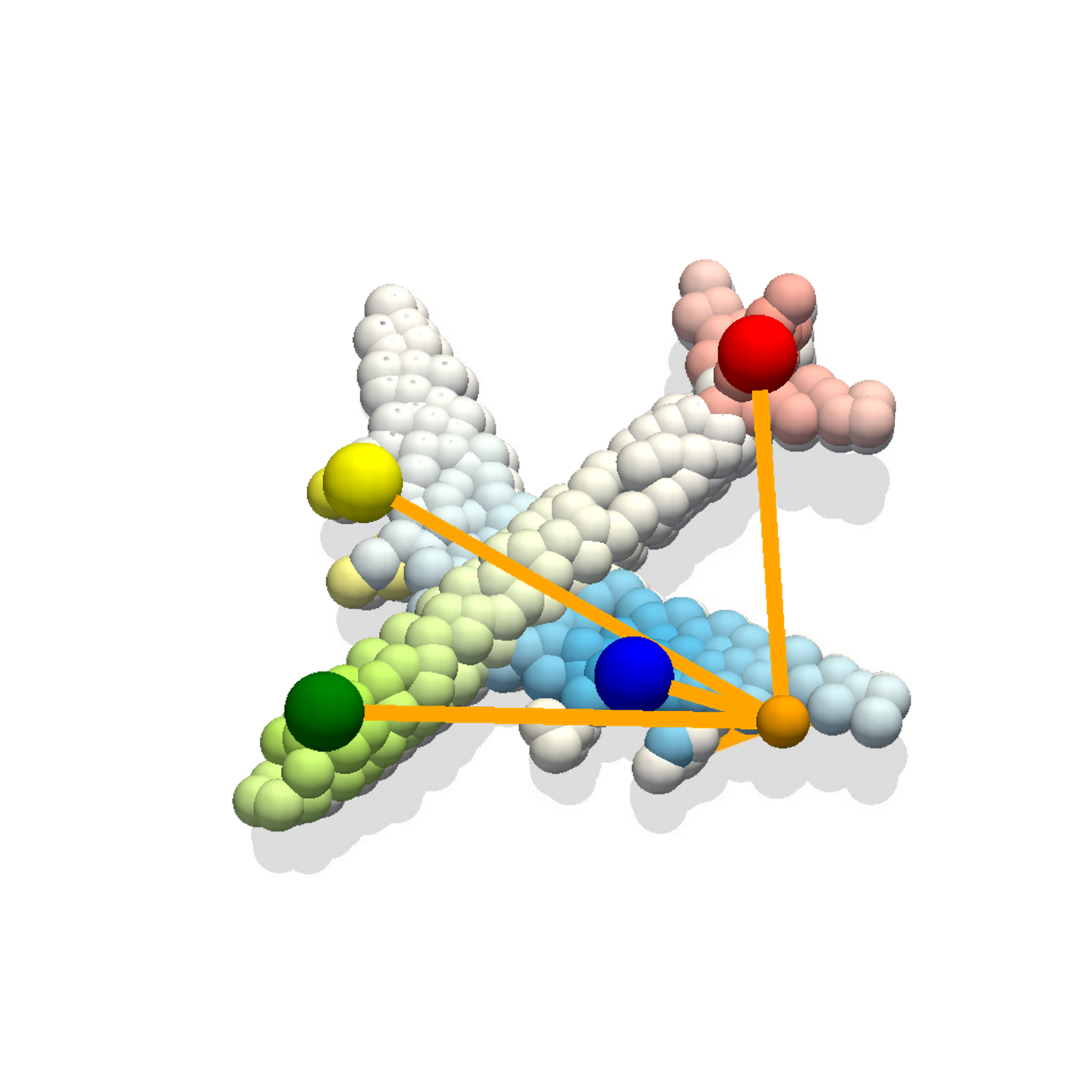}
         \caption{SP attention}
         \label{fig:spotr_att}
     \end{subfigure}
        \caption{{\textbf{Comparison of attention methods.} (a) Local attention, (b) Global attention, (c) Self-positioning point-based attention~(SP attention).}}
        \label{fig:attentions}

\end{figure}

Transformer~\cite{vaswani2017attention} tackled the long-range dependency issue in natural language processing and later it has been actively extended to 2D image processing~\cite{dosovitskiy2020image,liu2021swin,chu2021Twins,wang2021pyramid}.
Early works tried to replace convolutional layers with self-attention~\cite{dosovitskiy2020image,ramachandran2019stand,hu2019local,parmar2018image,chen2020generative,cordonnier2019relationship}, but they struggled with the quadratic computational cost of self-attention to the number of pixels.
To mitigate the scalability issue, self-attention in local neighborhoods~\cite{liu2021swin,wang2021pyramid} or approximating a self-attention with a reduced set of queries or keys~\cite{chu2021Twins,jaegle2021perceiver,zhu2020deformable} have been studied.
For point clouds, Point Transformer~\cite{zhao2021point} applies a local attention operation~(\Cref{fig:local_att}) and PointASNL \cite{yan2020pointasnl} employs a global attention module in a non-local manner~(\Cref{fig:global_att}).
Still, in point clouds, Transformer, which tackles both long-range dependency and scalability issues, has been less explored.

In this paper, we propose \textbf{S}elf-\textbf{Po}sitioning point-based \textbf{Tr}ansformer (SPoTr) to capture both local and global shape contexts with reduced complexity.
SPoTr block consists of two attention modules: (i) \textit{local points attention}~(LPA) to learn local structures and (ii) \textit{self-positioning point-based attention}~(SPA) to embrace global information via self-positioning points. 
SPA performs global attention by computing attention weights with only a small set of Self-Positioning points~(SP points) instead of the whole input points different from the standard global attention as illustrated in \Cref{fig:spotr_att}.
Specifically, SP points are adaptively located based on the input shape to cover the overall shape with only a small set of points. 
SP points learn its representation  considering both spatial and semantic proximity through \textit{disentangled attention}.
{Then, SPA non-locally distributes information of SP points to each input point.
We also show that our SPoTr block generalizes set abstraction~\cite{qi2017pointnet++} with improved expressive power.}

Further, we propose SPoTr architecture for standard point cloud tasks (\eg, shape classification and semantic segmentation).
We conduct extensive experiments with three datasets: ScanObjectNN~\cite{uy2019revisiting}, SN-Part~\cite{snpart}, and S3DIS~\cite{armeni20163d}.
Our proposed method shows its effectiveness on all datasets compared to other attention-based methods.
In particular, our architecture achieves an accuracy improvement of 2.6\% over the previous best model in shape classification with a real-world dataset ScanObjectNN.
Additionally, we demonstrate the effectiveness and interpretability of self-positioning point-based attention with qualitative analyses.  

The \textbf{contribution} of our paper can be summarized as the following:
\begin{itemize}
    \item[\textbullet] We design a novel Transformer architecture (SPoTr) to tackle the long-range dependency issues and the scalability issue of Transformer for point clouds.
    \item[\textbullet] We propose a global cross-attention mechanism with flexible self-positioning points (SPA). SPA aggregates information on a few self-positioning points via disentangled attention and non-locally distributes information to semantically related points.
    \item[\textbullet] SPoTr achieves the best performance on three point cloud benchmark datasets (SONN, SN-Part, and S3DIS) against strong baselines.
    \item[\textbullet] Our qualitative analyses show the effectiveness and interpretability of SPA.
\end{itemize}

\section{Related works}
    \label{sec:2}
\paragraph{Deep learning on point clouds.}
The success of CNNs has encouraged adapting CNNs to operate on point clouds rather than using hand-designed features.
Early approaches aim to transform the unstructured point cloud data into a structured form for directly applying convolution.
These include~\cite{su2015multi,guo2016multi,qi2016volumetric,jaritz2019multi,goyal2021revisiting,hamdi2021mvtn}, where they project 3D point clouds to 2D multi-view images for applying 2D convolution.
Other approaches~\cite{liu2019point,maturana2015voxnet,zhou2018voxelnet} convert point clouds to voxel grids, then apply 3D convolution.
However, both approaches have difficulty in preserving intrinsic geometries of point clouds.
To address this issue,  PointNet \cite{qi2017pointnet} directly processes point clouds with multi-layer perceptrons and a max-pooling function. 
However, it blindly aggregates all points without considering local information.
Thus, PointNet++~\cite{qi2017pointnet++} proposes utilizing local information through set abstraction and local grouping.
For further understanding of local contexts, recent works~\cite{atzmon2018point,li2018pointcnn,wu2019pointconv,thomas2019kpconv,wang2018deep,xu2021paconv,xu2018spidercnn,zhou2021adaptive,ma2022rethinking, ran2022surface} have proposed explicit convolution kernels on the point space.
KPConv~\cite{thomas2019kpconv} has applied deformable convolution~\cite{dai2017deformable,zhu2019deformable} to capture local information of point clouds.
PointNeXt~\cite{qian2022pointnext} has revisited PointNet++ by fully exploring its potential with improved training and augmentation schemes.
Although the representation power has been improved by capturing local information, the ability to capture long-range dependencies is limited.
SPoTr is the Transformer for point clouds equipped with global cross-attention to capture long-range dependencies.   

\paragraph{Attention-based methods on 2D images.}
Following the success of self-attention and Transformers~\cite{vaswani2017attention} in natural language understanding, many efforts have been made in the computer vision to replace convolution layers with self-attention layers~\cite{dosovitskiy2020image,ramachandran2019stand,hu2019local,parmar2018image,chen2020generative,cordonnier2019relationship}.
Despite the success, self-attention requires the quadratic computational cost with respect to the input image size.
To address the scalability issue, several works adopt self-attention within local neighborhoods~\cite{liu2021swin,wang2021pyramid}.
Swin Transformer~\cite{liu2021swin} utilizes non-overlapping windows and performs self-attention within each local window to get linear computational complexity in the number of input pixels.
Other works explore global attention mechanisms with a small set of queries or keys to reduce complexity~\cite{chu2021Twins,jaegle2021perceiver,zhu2020deformable}.
Twins~\cite{chu2021Twins} applies attention with a small set of representatives. 
Inspired by recent works, we suggest an efficient global cross-attention with only a small set of self-positioning points for point clouds.

\paragraph{Attention-based methods on point clouds.}
Recently, \cite{yan2020pointasnl,xie2018attentional,lee2019set,guo2021pct,zhao2021point,ran2021learning,mazur2021cloud,choe2022pointmixer, xiang2021walk} have adopted attention operations for point cloud processing. 
PointASNL~\cite{yan2020pointasnl} leverages the attention operation to non-locally influence entire points.
RPNet~\cite{ran2021learning} proposes attention-based modules for capturing local semantic and positional relations.
PointTransformer~\cite{zhao2021point} performs self-attention only within local neighborhoods.
CloudTransformer~\cite{mazur2021cloud} inspired by spatial transformer~\cite{jaderberg2015spatial}, uses an attention mechanism to transform the point cloud into a voxel grid for convolutional operation.
Although these works have proven to be effective, most works have neglected the capability of Transformers to capture long-range dependencies due to their quadratic computational cost to the number of input points.
In this paper, we aim to design Transformer architecture to capture both local and global information with a modest computational cost.

\section{Method}
    \label{sec:3}
\begin{figure*}[t] 
\centering
\includegraphics[width=1.0\textwidth]{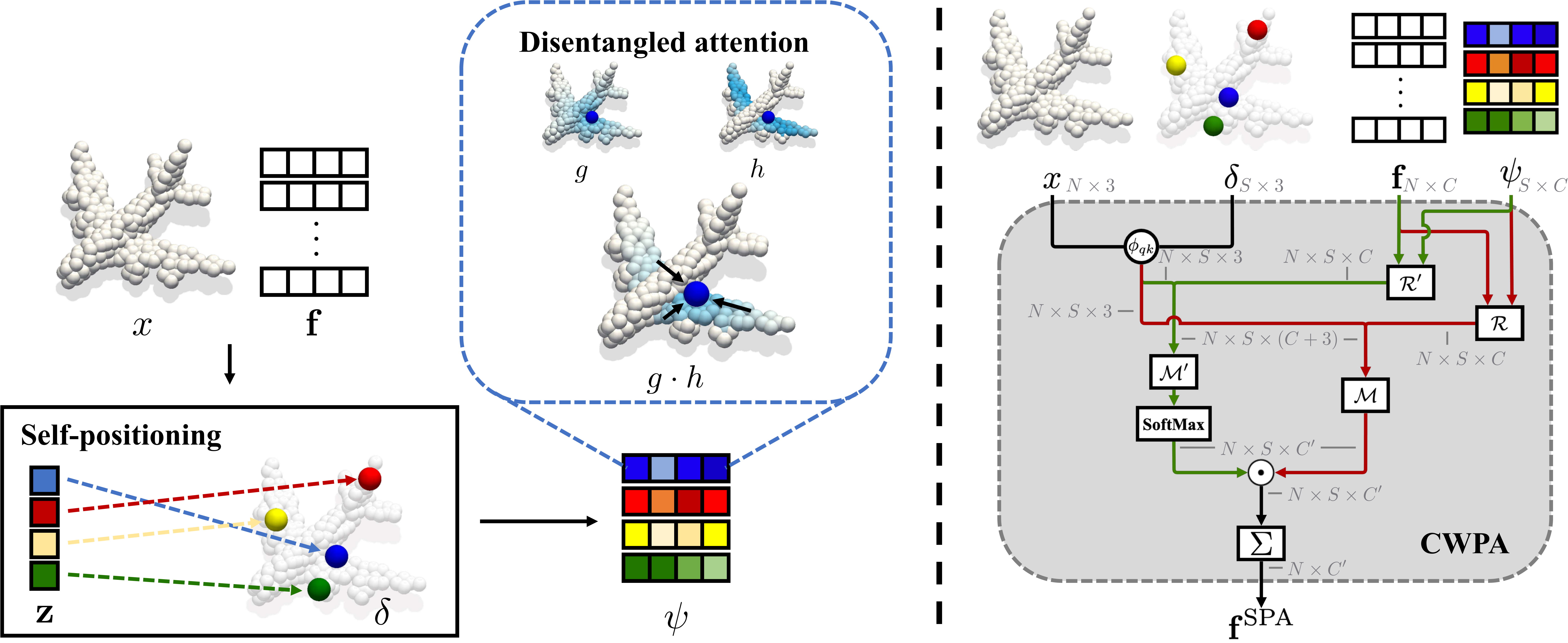}
\caption{{{\textbf{Illustration of self-positioning point-based attention (SPA).} Given input points $\Cen$ and their corresponding features $\Feat$, self-positioning points~(SP points) $\Vpoint$ are adaptively placed through the learnable latent $\Zfeat$. SP points aggregate features considering both spatial and semantic proximity and constructs $\Vfeat$ via disentangled attention. Then, SPA performs channel-wise point attention (CWPA) between input points and SP points to generate the output features $\Feat^{\text{SPA}}$.
}}}
\label{fig:fig2}
\end{figure*}

The goal of our framework SPoTr is to learn point representations for various point cloud processing tasks with a Transformer architecture using \textit{self-positioning points}.
First, we shortly describe the background regarding the point-based approaches including PointNet++ and Point Transformer~(\Cref{sec:3.1}). 
Second, we propose self-positioning point-based attention to efficiently capture the global context~(\Cref{sec:3.2}).
Third, we delineate the SPoTr block, which compromises both global cross-attention and local self-attention, and discuss the relation with a popular point-based network~(\Cref{sec:3.3}). 
Finally, we present the overall architecture of SPoTr, which is composed of multiple SPoTr blocks, for shape classification and segmentation tasks~(\Cref{sec:3.4}).



    \subsection{Backgrounds}
    \label{sec:3.1}


In this subsection, we briefly revisit the point-based approaches such as  PointNet++~\cite{qi2017pointnet++} and Point Transformer~\cite{zhao2021point}.

\paragraph{PointNet++~\cite{qi2017pointnet++}} captures local shape information through set abstraction and local grouping.
Given that a point set $\mathcal{P} = \left\{x_i \right\}_{i=1}^N$, where $x_i$ is the position of the $i$-th point, and its corresponding feature $\mathbf{f}_i$, PointNet++ proposed local set abstraction as follows:
\begin{equation}
    \mathbf{f}_i' = \mathcal{A}\left(\left\{\mathcal{M}\left([\mathbf{f}_j;\phi_{ij}] \right),\  \forall j \in \mathcal{G}_i \right\}\right),
\end{equation}
where $\mathcal{M}$ is the mapping function~(\eg, MLP), $\mathcal{A}$ is the aggregation function such as max-pooling, and $\mathcal{G}_i$ is the index set of the local group centered on the $i$-th point.

\paragraph{Point Transformer~\cite{zhao2021point}} leverages self-attention operations~\cite{zhao2020exploring} to represent local point groups.
Similar to PointNet++, Point Transformer leverages local grouping to represent local point groups with a self-attention mechanism as follows:
\begin{equation}
    \begin{split}
    &\mathbf{f}_i^\prime = \sum_{j \in \Gc_i} \mathbf{A}_{ij} \odot\left(\mathbf{W}_1\mathbf{f}_j + \Delta_{ij} \right), \\
    &
    \mathbf{A}_{ij} = \text{SoftMax}\left(\mathcal{M}\left(\mathbf{W}_2\mathbf{f}_i - \mathbf{W}_3\mathbf{f}_j + \Delta_{ij} \right)\right),  
    \end{split}
\end{equation}
where $\odot$ denotes element-wise multiplication,  $\mathbf{W}_1$, $\mathbf{W}_2$, $\mathbf{W}_3$ are learnable weight matrices, $\mathcal{M}$ is a mapping function such as multi-layer perceptron, and $\Delta$ is a positional encoding.
Point Transformer~\cite{zhao2021point} has shown the advantage of the attention mechanism on point clouds only with `local attention' since computing the global attention on whole input points is almost infeasible on large-scale data.

    \subsection{Self-positioning point-based attention}
\label{sec:3.2}


We propose an efficient global attention called \textit{self-positioning point-based attention} (SPA) to resolve the inherent limitation of Transformer for point clouds (\eg, scalability) as illustrated in~\Cref{fig:fig2}.
Herein, SPA computes attention weights with only a small set of self-positioning points named \textit{SP points}.
Since the overall shape context is captured by only a small number of SP points, the position of the SP points $\Vpoint_s$ needs to be \textit{flexible}, so that the points become sufficiently representative of local part shapes.
With our method, the SP points are adaptively located according to the input shape (see~\Cref{sec:4.3} for visualizations).
Similar to the offsets in deformable convolutional neural networks~\cite{dai2017deformable} calculated with the feature of each pixel, we calculate $\Vpoint_s$ with its latent vector $\Zfeat_s$ and feature $\Feat_i$ of all input points $\forall x_i \in \mathcal{P}$, where $\Pc$ is the set of points.
We adopt adaptive interpolation and compute the positions of SP points as
\begin{equation}
\label{eq:spgp}
    \Vpoint_s = \sum_{i} \text{SoftMax}\left(\Feat_i^\top \Zfeat_s\right)\Cen_i.
\end{equation}
Hence, SP points are always located nearby input points and precisely they stay within the convex hull of input points.

Then, SPA, equipped with SP points, performs global cross-attention in two steps: aggregation and distribution.
At the aggregation step, SPA aggregates the features from all input points considering both spatial and semantic proximity. It can be written as
\begin{equation}
\label{eq:bilateral}
    \Vfeat_s = \sum_i g\left(\Vpoint_s, \Cen_i  \right)\cdot h\left(\Zfeat_s, \Feat_i\right)\cdot \Feat_i,
\end{equation}
where $g, h$ are spatial and semantic kernel functions, respectively.
For the spatial kernel function $g$, we use the Radial Basis Function~(RBF) as
\begin{equation}
\label{eq:spatial}
    g\left(\Vpoint_s, \Cen_i\right) = \sum_i \exp\left(-\gamma \left\lVert \Vpoint_s - \Cen_i\right\rVert^2 \right),
\end{equation}
where $\gamma \in \mathbb{R}_{++}$ is a bandwidth that adjusts the size of receptive fields.
If $\gamma$ has a higher value, the size of receptive fields gets smaller.
For the semantic kernel function $h$, we utilize the attention-based kernel as below:
\begin{equation}
\label{eq:semantic}
    h\left(\Zfeat_s, \Feat_i\right) = \frac{\exp\left(\Feat_i^\top \Zfeat_s\right)}{\sum_{i^\prime}\exp\left(\Feat_{i^\prime}^\top \Zfeat_s\right)}.
\end{equation}
Only considering the spatial information can cause \textit{information smoothing}, \ie, the information from neighboring points with different semantics can reduce the descriptive power.
Thus, we consider both spatial proximity and semantic proximity through two separate kernels $h$ and $g$.
These separate kernels can be interpreted as \textit{disentangled attention}.
The disentangled attention, which works similarly to the bilateral filter, allows the SP points to have greater descriptive power.
Further analyses on how it processes, can be seen in ~\Cref{sec:4.3}. 
Finally, at the distribution step, SPA performs the cross-attention between SP points and input points as follows:
\begin{equation}
\label{eq:globalatt}
    \Feat_i^\text{SPA} = \mbox{CWPA}\left(\Cen_i, \Feat_i, \left\{\Vpoint_s\right\}_{s=1}^S, \left\{\Vfeat_s\right\}_{s=1}^S\right),
\end{equation}
where $\Feat_i^\text{SPA}$ is the final output of the SPA, $\mbox{CWPA}$ is channel-wise point attention. CWPA will be described next.


\paragraph{Channel-wise Point Attention.}
Channel-Wise Point Attention~(CWPA) computes attention weight between query and key points for each channel, different from the standard attention that generates the same attention weight across channels.
CWPA is formulated as follows:
\begin{equation}
\begin{split}
    &\mbox{CWPA}\left(x_{q}, \Feat_q, \left\{x_k\right\}_{k \in \Omega_{\text{key}}}, \left\{\Feat_k\right\}_{k \in \Omega_{\text{key}}}\right) \\
    &=  \sum_{k \in \Omega_{\text{key}}} \mathbb{A}_{q,k,:} \odot \mathcal{M}\left([\mathcal{R}(\mathbf{f}_q,\mathbf{f}_k);\phi_{qk}] \right),
\end{split}
\end{equation}
where $x_q, x_k \in \mathbb{R}^3$ are the positions of the query/key points and $\Feat_q, \Feat_k \in \mathbb{R}^C$ are their corresponding features. $\Omega_{key}$ denotes the set of key points. 
To take into account the relative information of contexts and positions, we use $\phi_{qk}$ and $\mathcal{R}(\cdot)$.
$\phi_{qk}$ is a normalized relative position of the key point based on the query point, $\mathcal{R}$ is the relation function between query and key feature~(\eg, $\mathbf{f}_q-\mathbf{f}_k$) and $\mathcal{M}$ indicates the mapping function.
The channel-wise point attention $\mathbb{A}_{q,k,:} \in \mathbb{R}^C$ between the query and key point is defined as follows:
  \begin{equation}
     \mathbb{A}_{q,k,c} = \frac{\exp\left(\mathcal{M}'\left([\mathcal{R}'(\mathbf{f}_q,\mathbf{f}_k);\phi_{qk}]/\tau \right)_c \right)}{\sum_{k' \in \Omega_{key}}\exp\left(\mathcal{M}'\left([\mathcal{R}'(\mathbf{f}_q,\mathbf{f}_{k'});\phi_{qk'}]/\tau \right)_c \right)},
 \end{equation}
where c is the index of channel, and $\tau$ denotes temperature.
$\mathcal{M}^\prime$ and $\mathcal{R}^\prime$ are the mapping function and the relation function, respectively.
In our implementation, we use $\mathcal{R}^\prime = \mathbf{f}_q - \mathbf{f}_k$ same as $\mathcal{R}$.
By adopting the proposed channel-wise point attention, CWPA can learn more powerful and flexible representations compared to standard attention.


    \subsection{Self-positioning point-based Transformer}
    \label{sec:3.3}

\begin{figure*}[ht] 
\centering
\includegraphics[width=1\textwidth]{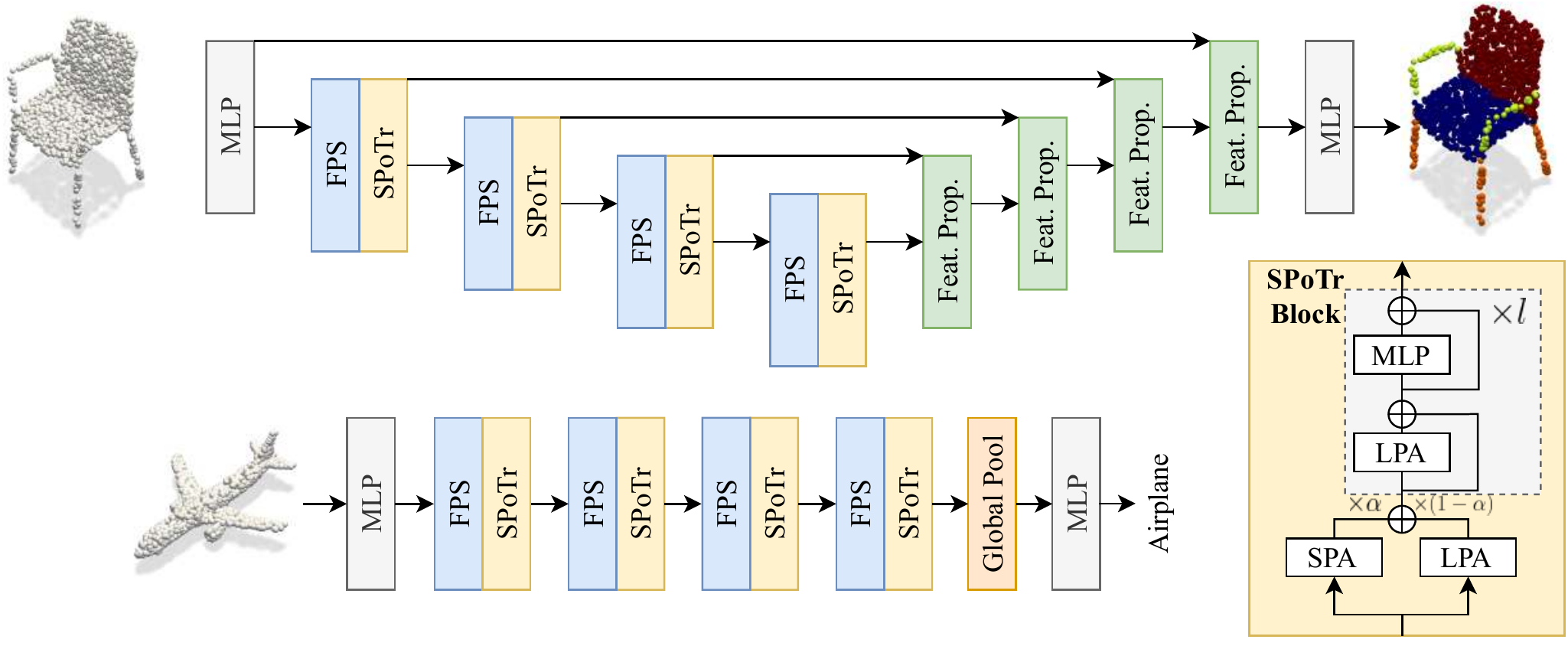}
\caption{
\textbf{Overall Architecture of SPoTr.} For classification~(bottom), four SPoTr blocks run consecutively, followed by a max-pooling and a multi-layer perceptron. 
For segmentation~(top), a U-net style architecture is adopted with SPoTr blocks for downsampling and feature propagation for upsampling, followed by a multi-layer perceptron.}
\label{fig:fig4}
\end{figure*}

We now propose the SPoTr block that utilizes \emph{self-positioning point-based attention}~(SPA) with \emph{local point attention}~(LPA).
By combining LPA and SPA, it captures not only local and short-distance information but also long-distance and global information.

\paragraph{Local points attention (LPA).} 
We adopt local points attention~(LPA) defined on a local group to learn local shape context.
A local point group consists of neighbors in ball query centered on an anchor point $x_i$.
The attention for each local point group $\Pointgroup_i$ and points $\left\{x_j | \forall j \in \Pointgroup_i\right\}$ is defined as
\begin{equation}
\label{eq:lpa}
    \Feat_{i}^{\text{LPA}} = \mbox{CWPA}\left(x_{i}, \Feat_i, \left\{x_j\right\}_{j \in \Pointgroup_i}, \left\{\Feat_j\right\}_{j \in \Pointgroup_i}\right),
\end{equation}
where $\Feat_{j}$ is the feature vector of point $x_j$, $\Feat^\text{LPA}_{i}$ is an output feature vector of LPA.
We adopt channel-wise point attention operation~(CWPA) same as SPA.

\paragraph{SPoTr block.}
We construct a SPoTr block by combining the local points attention~(LPA) module and self-positioning point-based attention (SPA) module to capture local and global information simultaneously.
As shown in \Cref{fig:fig4}~(bottom right), the SPoTr block is defined as follows:
\begin{equation}
    \hat{\Feat}_i = \alpha \cdot \Feat_i^{\text{SPA}} + (1-\alpha) \cdot \Feat_i^{\text{LPA}} 
\end{equation}
where $\alpha$ is a learnable parameter that softly selects the representations generated by self-positioning point-based attention and local points attention.
Finally, LPA and MLP with batch normalization and the residual connection are applied to extract point-wise high-level representations.

\paragraph{Connection between SPoTr block and Set abstraction.}
We show the superiority of the SPoTr block by discussing it with set abstraction in PointNet++~\cite{qi2017pointnet++}. 
\begin{remark}
A SPoTr block with proper $\alpha, \mathcal{M},\mathcal{M}',\mathcal{R},\mathcal{R}'$ can express set abstraction.
\end{remark}
When the value of $\tau$ is sufficiently low, the function $\mathcal{M}^\prime$ is the same as $\mathcal{M}$, $\alpha=0$, and $\mathcal{R}(\mathbf{f}_q, \mathbf{f}_k)=\mathcal{R}^\prime(\mathbf{f}_q, \mathbf{f}_k)=\mathbf{f}_k$, the channel-wise point attention becomes equivalent to the set abstraction.   
In this setting, the attention weight between the query point and $k$-th key point on $c$-th channel becomes 1 if $k = \underset{k^\prime \in \Omega_k}{\mathrm{argmax}}\  \mathcal{M}\left([\mathbf{f}_{k^\prime};\phi_{qk^\prime}] \right)_{q,k^\prime,c}$.
Otherwise, the attention score is 0.
It means that the attention only activates the maximum channels alike a max-pooling operation. 
Therefore, the SPoTr block can play a role as a max-pooling operation following the mapping function, which is the set abstraction.
This fact supports the improved expressive power of SPoTr on point cloud analysis.

    \subsection{SPoTr architectures}
    \label{sec:3.4}

We design a Transformer-based architecture called SPoTr for point cloud tasks as illustrated in~\Cref{fig:fig4}. 

\paragraph{Classification.}
For the shape classification task, we build our Transformer encoder by stacking the SPoTr blocks  described in \Cref{sec:3.3}.
To increase the representational power, we first apply an MLP before operating the attention blocks following \cite{zhao2021point}.
Then, the SPoTr blocks are sequentially applied on sampled points, which are sampled through the farthest point sampling~(FPS).
In shape classification, we set $l=0$ for the SPoTr block since we empirically found that it is enough in shape classification task.
Besides, the sampling rates are 1/4 for every stage.
We use the ball query that selects points within a radius (an upper limit of the number of neighborhoods is set in implementation) to generate a local point group $\Pointgroup_i$ centered at point $x_i$ following \cite{qi2017pointnet++}.
After the last stage, the features are aggregated by a max-pooling function and processed by an MLP.

\paragraph{Segmentation.}
The encoder for semantic segmentation contains the SPoTr blocks and FPS.
Following previous studies~\cite{qi2017pointnet++}, we apply a U-net designed architecture, which contains the feature propagation layers and the SPoTr blocks for dense prediction.
Same to the classification, we use a ball query to generate a local group.
The outputs of the final block are processed by an MLP.
More details on SPoTr architectures are in the supplement.

\section{Experiments}
    \label{sec:4}
In this section, we demonstrate the effectiveness of SPoTr and provide quantitative and qualitative results for further analyses.
First, we conduct shape classification and semantic segmentation~(\Cref{sec:4.1}).
We also provide ablation studies and complexity analysis of SPoTr~(\Cref{sec:4.2}).
Lastly, we provide visualizations to better understand how SPA behaves~(\Cref{sec:4.3}). Implementation details are available in the supplement.

    \subsection{Shape classification and semantic segmentation}
    \label{sec:4.1}

\begin{table}[t]
  \centering
\setlength{\tabcolsep}{8pt}
  \begin{tabular}{l|c|c c}
    \toprule
    Methods & Year & mAcc & OA \\
    \midrule
    PointNet~\cite{qi2017pointnet} & 2017 &  63.4 &68.2\\
    PointNet++~\cite{qi2017pointnet++} & 2017 & 75.4 & 77.9 \\
    SpiderCNN~\cite{xu2018spidercnn}& 2018& 69.8 & 73.7\\
    PointCNN~\cite{li2018pointcnn}&2018& 75.1 & 78.5 \\
    DGCNN~\cite{wang2019dynamic}&2019& 73.6 & 78.1\\
    DRNet~\cite{qiu2021dense}&2021& 78.0 & 80.3 \\
    GBNet~\cite{qiu2021geometric}&2021& 77.8 & 80.5 \\
    SimpleView~\cite{goyal2021revisiting}&2021 &-& 80.5\\
    PRA-Net~\cite{cheng2021net}&2021& 77.9 & 81.0 \\
    MVTN~\cite{hamdi2021mvtn}&2021 & - & 82.8 \\
    CT~\cite{mazur2021cloud} & 2021 & 83.1 & 85.5 \\
    PointMLP~\cite{ma2022rethinking} & 2022 & 84.4 & 85.7 \\
    RepSurf-U~\cite{ran2022surface} & 2022 & 83.1 & 86.0 \\
    \rowcolor{LightYellow}PointNeXt~\cite{qian2022pointnext}& 2022 & 85.8±0.6 & 87.7±0.4\\
    \midrule
    \rowcolor{LightRed}\textbf{SPoTr}& 2023 & \textbf{86.8} &\textbf{88.6} \\
    \bottomrule
  \end{tabular}
  \caption{\textbf{Shape classification results on PB\_T50\_RS in SONN.}
  \label{tab:sonn}
  mAcc is the mean of class accuracy and OA is the overall accuracy.
  }
\end{table} 

\paragraph{Shape Classification.}
For the shape classification, we validate SPoTr on a real-world dataset ScanObjectNN (\textbf{SONN})~\cite{uy2019revisiting}. 
SONN has 2,902 objects categorized into 15 classes from SceneNN~\cite{hua2016scenenn} and ScanNet~\cite{dai2017scannet}.
Among diverse variants of SONN, we use PB\_T50\_RS (\textbf{SONN\_PB}), which is the most challenging version with random perturbation and contains 14,510 objects in total.
We follow the official split of \cite{uy2019revisiting}, where they divide SONN into 80\% for training and 20\% for evaluation.
Also, we sample 1,024 points for training and evaluating the models. 

\Cref{tab:sonn} shows that SPoTr outperforms all baselines with the mean of class accuracy (mAcc) of 86.8\% and overall accuracy (OA) of 88.6\% (+1.0\% mAcc, +0.9\% OA).
This result shows that capturing long-range context is important for recognizing 3D shapes in real-world datasets.
\begin{table}[t]
  \centering
\setlength{\tabcolsep}{5pt}
  \begin{tabular}{l|c|cc}
    \toprule
    Methods & Year & cls. mIoU & ins. mIoU\\
    \midrule
    PointNet~\cite{qi2017pointnet}&2017 & 80.4& 83.7\\
    PointNet++~\cite{qi2017pointnet++}&2017 & 81.9 & 85.1 \\
    PointCNN~\cite{li2018pointcnn}&2018& 84.6 & 86.1 \\
    DGCNN~\cite{wang2019dynamic}&2019 & 82.3 & 85.1 \\
    RSCNN~\cite{liu2019relation} & 2019 & 84.0 & 86.2 \\
    KPConv~\cite{thomas2019kpconv}&2019 & 85.1&86.4 \\
    PointConv~\cite{wu2019pointconv}&2019 & 82.8 & 85.7 \\
    PointASNL~\cite{yan2020pointasnl}&2020 & - & 86.1 \\
    PCT~\cite{guo2021pct}&2021 & - & 86.4\\
    PAConv~\cite{xu2021paconv}&2021 & 84.6  &  86.1 \\
    AdaptConv~\cite{zhou2021adaptive}&2021 & 83.4 & 86.4\\
    PointTransformer~\cite{zhao2021point}&2021& 83.7 & 86.6 \\
    CurveNet~\cite{xiang2021walk}&2021 & - & 86.8 \\
    PointMLP~\cite{ma2022rethinking} & 2022 &84.6 & 86.1\\
    \rowcolor{LightYellow}PointNeXt~\cite{qian2022pointnext} & 2022& 85.2 $\pm$ 0.1 & 87.0 $\pm$ 0.1 \\
    \midrule
    \rowcolor{LightRed}\textbf{SPoTr} & 2023& \textbf{85.4} & \textbf{87.2}\\

    \bottomrule
  \end{tabular}
  \caption{\textbf{Part segmentation results on SN-Part.} ins. mIoU is the mean of instance IoU. cls. mIoU is the mean of the class IoU. 
  }
  \label{tab:shapenet}
\end{table} 

\begin{table}[t]
  \centering
\setlength{\tabcolsep}{3pt}
\resizebox{\columnwidth}{!}{
  \begin{tabular}{l|c|ccc}
    \toprule
    Methods & Year & OA & mAcc & mIoU\\
    \midrule
    PointNet~\cite{qi2017pointnet} & 2017  & - & - & 41.1 \\
    PointCNN~\cite{li2018pointcnn} & 2018 &  85.9 & 63.9 & 57.3\\
    PointWeb~\cite{zhao2019pointweb} & 2019 & 87.0 & 66.6 & 60.3\\
    KPConv~\cite{thomas2019kpconv} & 2019 & - & 72.8 & 67.1 \\
    PCT~\cite{guo2021pct} & 2021 & - & 67.7 & 61.3 \\
    CT~\cite{mazur2021cloud} & 2021 & - & - & 67.9 \\
    PointTransformer~\cite{zhao2021point} & 2021 & \textbf{90.8} & - & 70.4 \\
    RepSurf-U~\cite{ran2022surface} & 2022 & 90.2 & 76.0 & 68.9 \\
    \rowcolor{LightYellow}PointNeXt~\cite{qian2022pointnext} & 2022 & 90.6 $\pm$ 0.1& - & 70.5 $\pm$ 0.3 \\
    \midrule
    \rowcolor{LightRed}\textbf{SPoTr}  & 2023 & 90.7 & \textbf{76.4} & \textbf{70.8}\\ 
    \bottomrule
  \end{tabular}}
  \caption{\textbf{Semantic segmentation results on S3DIS.} OA is the overall accuracy, mAcc is the mean of class accuracy, and mIoU is the mean of instance IoU.
  \label{tab:s3dis}
  }

\end{table}

\paragraph{Part Segmentation.}
For part segmentation, we use \textbf{SN-Part}~\cite{snpart,lee2022sagemix}, which is a synthetic dataset with 16,881 shapes from 16 categories with 50 part labels.
We follow the split used in \cite{qi2017pointnet}, where 14,006 samples are for training and 2,874 samples are for validation. 
On each shape, 2,048 points are randomly sampled.

The results are reported in~\Cref{tab:shapenet}, where we evaluate the performance with the mean of instance IoU (ins. mIoU) and class IoU (cls. mIoU). Following previous works~\cite{liu2019relation,xiang2021walk,xu2021paconv}, we report the results with a multi-scale inference setting. 
Although the performance in SN-Part is quite saturated, SPoTr achieves the best performance 87.2\% with considerable improvements (+0.2\% mIoU).

\paragraph{Scene Segmentation.}
For comparison with previous methods~\cite{qi2017pointnet,thomas2019kpconv,zhao2021point,choe2022pointmixer} on scene segmentation, we validate SPoTr on the widely used benchmark dataset \textbf{S3DIS}~\cite{armeni20163d}. S3DIS is the large-scale dataset containing 271 rooms from 6 indoor areas with 13 semantic categories. In our experiments, we largely follow the settings of PointTransformer~\cite{zhao2021point} and consider Area-5 as the test set.

As shown in~\Cref{tab:s3dis}, SPoTr outperforms all previous methods in every metric (\ie, overall accuracy (OA), mean of class accuracy (mAcc), and mean of instance IoU (mIoU)).
The superior performance over previous Transformer architecture~\cite{zhao2021point} (+0.4\% mIoU) proves the importance of long-range dependency in the semantic segmentation as well as the shape classification.
    \subsection{Quantitative analysis}
    \label{sec:4.2} 

\begin{table}[t]
  \centering
\setlength{\tabcolsep}{5pt}
  \begin{tabular}{l|ccc|cc}
    \toprule
    Method & $g$ & $h$ & SP&  OA\\
    \midrule
    w/o SPA (\textit{baseline}) &    &  &   &  87.9\\
    w/o self-positioning &  \checkmark  & \checkmark &&87.7\\
    w/o disentangled attention &\checkmark &  &  \checkmark  &88.2\\
    SPoTr (\textit{ours}) & \checkmark &\checkmark & \checkmark &\textbf{88.6}\\
    \bottomrule
  \end{tabular}
  
    \caption{\textbf{Ablations on SONN\_PB.} $g$: spatial kernel, $h$: semantic kernel, SP: self-positioning points. OA is the overall accuracy.}
\label{table5}
\end{table}

\begin{table}[t]
  \centering
\setlength{\tabcolsep}{5pt}
  \begin{tabular}{l|c|c}
    \toprule
    Attention type& Semantic rel. $\mathcal{R}$ &   OA\\
    \midrule
    Standard Att.  &-- & 86.1\\
    CWPA &$\mathbf{f}_k$    & 88.1\\
    CWPA & $\mathbf{f}_q + \mathbf{f}_k$    & 86.4\\
    CWPA & $\mathbf{f}_q \odot \mathbf{f}_k$&  85.4\\
    CWPA &$\mathbf{f}_q - \mathbf{f}_k$ & \textbf{88.6}\\
    \bottomrule
  \end{tabular}
  
    \caption{\textbf{Performance comparisons of different attention types and semantic relation $\mathcal{R}$ on SONN\_PB.} Attention types : Standard Attention in Transformer~\cite{vaswani2017attention} and channel-wise point attention~(CWPA) with Semantic relation : $\mathcal{R}(\mathbf{f}_q, \mathbf{f}_k)$} 
  \label{table6}
\end{table} 

\begin{figure*}[ht] 
\centering
\includegraphics[trim=20 20 20 20,clip,width=1.0\textwidth]{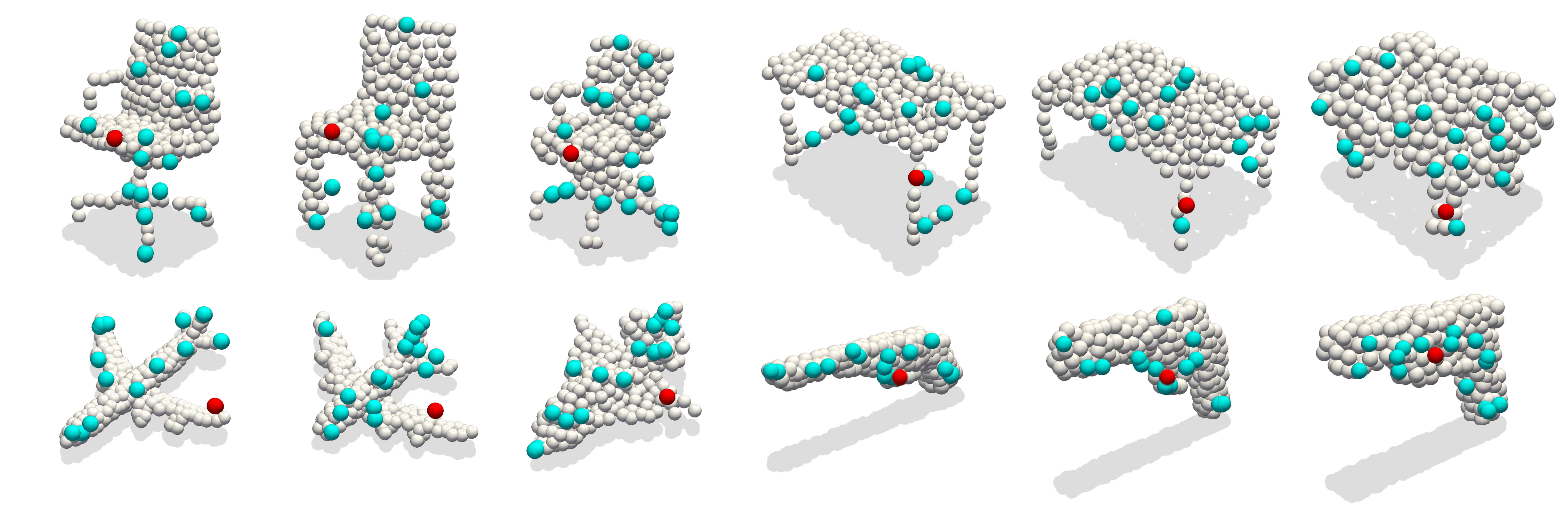}
\caption{{\textbf{Self-positioning points~(SP points).} \textcolor{cyan}{SP points} are adaptively self-positioned according to each shape. \textcolor[rgb]{1,0,0}{Red points} correspond to specific SP points. Under the same class, the red points are located at \textit{semantically similar} positions.
}}
\label{fig:SPpoints}
\end{figure*}
\begin{figure}[t] 
\centering

\includegraphics[trim=50 20 30 10, width=1.0\columnwidth]{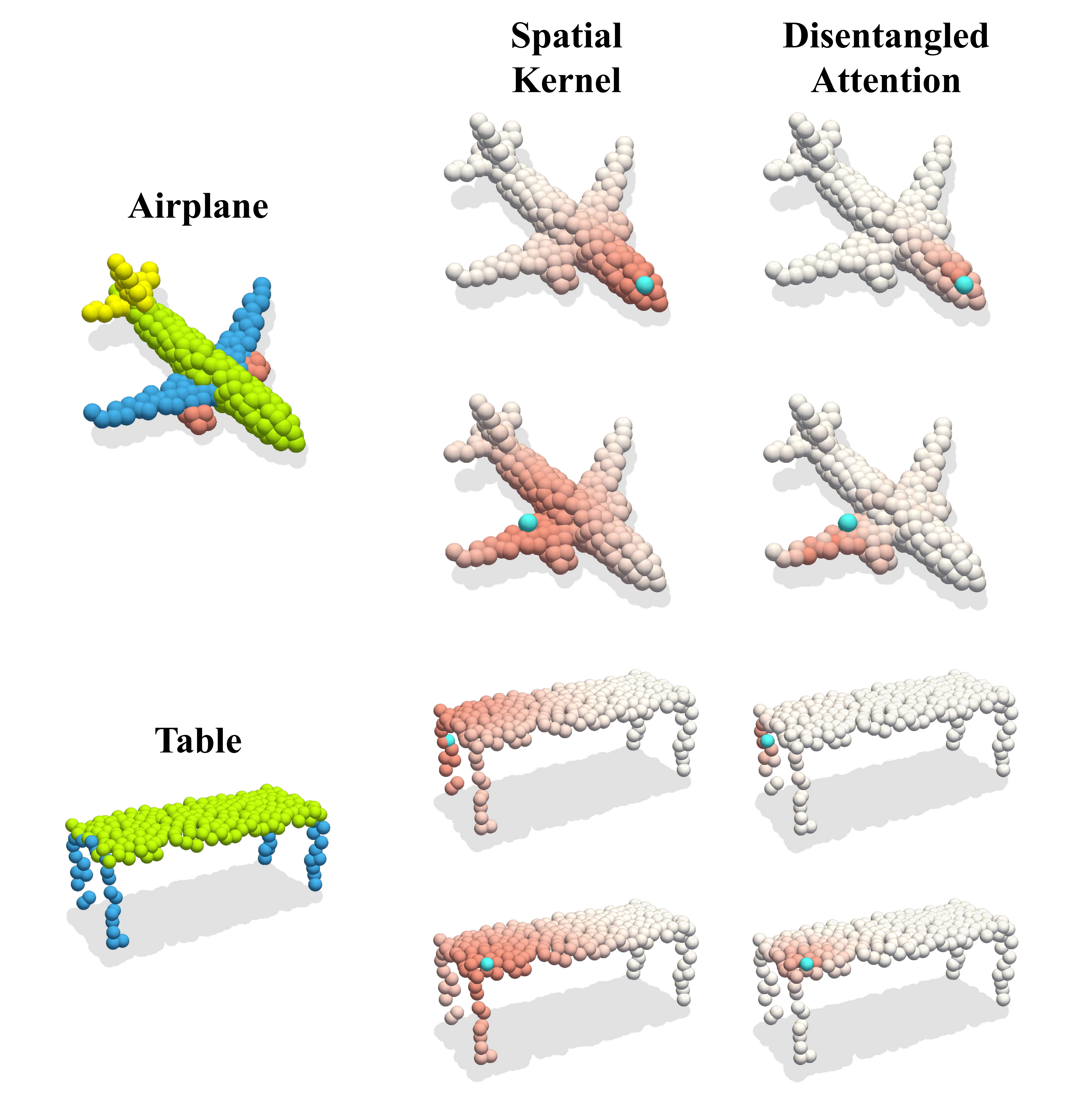}
\caption{
\textbf{Visual comparison of disentangled attention with a spatial kernel.}
{
A spatial kernel (Middle) only considers spatial proximity without considering semantic relevance. Differently, Disentangled Attention~(Right) filters out irrelevant information.}}

\label{fig:dis_att}
\end{figure}


\paragraph{Ablation studies.}
We explore how self-positioning positions (SP) and disentangled attention contribute to SPA.
\Cref{table5} shows the final results on SONN, where the baseline~(\textit{w/o SPA}) learns only with local point attention.
In the case of \textit{w/o self-positioning}, we use FPS to randomly select a small set of points for cross-attention, and for \textit{w/o disentangled attention}, we only adopt the spatial kernel function $g$.
Our model with all the components of SPA achieves the best performance of 88.6\% in overall accuracy.
This superior performance verifies that every component is crucial for SPA.
In particular, when we use FPS instead of SP, the performance is even worse than the baseline as overall accuracy dropped from 87.9\% to 87.7\%.
This observation suggests the positions of SP points \textit{matter} for global cross-attention.
Rather than simple sampling, our learnable approach successfully locates SP points and makes global cross-attention effective.
Next, with \textit{w/o disentangled attention}, the performance gain in OA is minimal (0.3\%) over the baseline compared to using disentangled attention (0.7\%).
It indicates that disentangled attention improves the descriptive power by filtering semantically irrelevant information.

\paragraph{Attention types and semantic relation $\mathcal{R}$.}
In \Cref{table6}, we conduct experiments to compare the models with different attention types (Standard attention in Transformer~\cite{vaswani2017attention} and our CWPA) and semantic relations ($\mathcal{R}(\mathbf{f}_q, \mathbf{f}_k) = $ $\mathbf{f}_k$, $\mathbf{f}_q+\mathbf{f}_k$, $\mathbf{f}_q\odot\mathbf{f}_k$, and $\mathbf{f}_q-\mathbf{f}_k$).
The models adopting the CWPA outperform the model with the standard attention, which shows that the channel-wise point attention operation is more powerful to represent point clouds compared to the standard attention.
Furthermore, the results demonstrate that Sub~($\mathbf{f}_q-\mathbf{f}_k$) is most appropriate to model the semantic relation between points. 

\begin{table}[t]
  \centering
\resizebox{\columnwidth}{!}{
  \begin{tabular}{l|cccc}
    \toprule
    Method & Param $\downarrow$ & FLOPs $\downarrow$ &  Memory $\downarrow$& Throughput $\uparrow$\\
    &(M)&(G)&(GB)&(shapes/s)\\
    \midrule
    GSA & \textbf{1.7} & 114.0 & 24.2 & 17.7 \\
    SPA (\textit{ours}) & \textbf{1.7} & \textbf{10.8} &\textbf{ 2.5}& \textbf{281.5} \\
    & (-)  & (- 90.5\%)  &   (- 89.7\%)  &  ($\times$ 15.9) \\
    \bottomrule
  \end{tabular}}

\caption{\textbf{Complexity analysis on SN-Part.} SPA: self-positioning point-based attention, GSA: global self-attention.}
\label{tab:complexity}
\end{table} 

\paragraph{Complexity analysis on SN-Part.} 
We analyze the space and time complexity to validate the computational efficiency of SPoTr during inference time with a batch size of 8.
For a baseline, SPA in SPoTr is replaced by the standard global self-attention (abbreviated in GSA) with CWPA. 
For a comparison with GSA requiring the quadratic complexity, we inevitably use the variants of SPoTr, where the channel size of each layer is reduced by $\times 1/4$.
For space complexity, we measure  the number of parameters and total memory usage, and for time complexity, we measure FLOPs and throughput performance.
\Cref{tab:complexity} empirically proves the efficiency over GSA.
For space complexity, GSA shares a similar number of parameters with SPA but introduces a large memory usage of 24.2 (GB). Instead, Our SPA only uses 2.5 (GB) (-89.7\%). Also, SPA largely reduces the time complexity from 114.0 GFLOPS with a throughput of 17.7 (shapes/s) to 10.8 GFLOPS (-90.5\%) with a throughput of 281.5 (shapes/s) ($\times$15.9).

    \subsection{Qualitative analysis}
    \label{sec:4.3}
For a deeper understanding of each component in SPA, such as self-positioning points~(SP points) and disentangled attention, we provide qualitative results in this section.
We use SN-Part for visualizations.

\paragraph{Self-positioning points.}
As mentioned in Section~\ref{sec:3.2}, it is important that SP points are adaptively located considering the input shape.
\Cref{fig:SPpoints} shows that the SP points are adaptively located on various samples from different categories.
Interestingly, a specific SP point (colored in red) appears at a \textit{semantically similar} place for each category.
For example, red dots from airplanes are always near the left-wing.
This consistent placement of SP points implies that each SP point learns to represent semantically similar regions.

\paragraph{Disentangled attention.}
SPA aggregates feature considering spatial proximity as well as semantic proximity via disentangled attention as introduced in \Cref{sec:3.2}. 
\Cref{fig:dis_att} shows the weights of the spatial kernel $g$ and the effective receptive field of the disentangled attention $g\cdot h$.
Cyan-colored points are selected SP points and kernel weights are illustrated in heatmaps.
With only the spatial kernel $g$, SP point blindly aggregates the information of neighbor points inducing \textit{irrelevant} information from close regions (\eg, a wing and a body of an airplane are strongly colored in the second row of the figure).
Conversely, with our disentangled attention $g\cdot h$, the same SP point selectively aggregates information considering both spatial proximity and semantic proximity~(\eg, the right-wing is only colored in the figure).
The obvious difference suggests disentangled attention is crucial for enhancing the descriptive power by suppressing irrelevant information.

\section{Conclusion}
    \label{sec:5}
In this paper, we propose SPoTr, a Transformer for point clouds, which captures both local and global shape context without the quadratic complexity of input points.
SPoTr includes two attention modules: self-positioning point-based attention~(SPA) and local points attention~(LPA).
SPA is a novel global cross-attention, which aggregates information via disentangled attention and non-locally distributes information to entire points.
Our experiments show superior performance across various tasks, including ScanObjectNN, SN-Part, and S3DIS.

\paragraph{Acknowledgements.}
This work was supported by the MSIT, Korea, under the ICT Creative Consilience program (IITP-2023-2020-0-01819) supervised by the IITP and the Virtual Engineering Platform Project  (P0022336) of the Ministry of Trade, Industry and Energy (MOTIE), Korea.

{\small
\bibliographystyle{unsrt}
\bibliography{egbib}
}

\end{document}